\pdfoutput=1

\documentclass[11pt]{article}

\usepackage[]{EMNLP2023}

\usepackage{times}
\usepackage{latexsym}
\usepackage{graphicx}
\usepackage{float}
\usepackage{multirow}
\usepackage{booktabs}
\usepackage{amsmath,amsfonts}
\usepackage[T1]{fontenc}

\usepackage[utf8]{inputenc}

\usepackage{microtype}

\usepackage{inconsolata}

\title{Document Set Expansion with Positive-Unlabeled Learning: A Density Estimation-based Approach }

\author{ Haiyang Zhang$^{1*}$, Qiuyi Chen$^{1}$, Yuanjie Zou$^{1}$, Yushan Pan$^{1}$, Jia Wang$^{1}$, Mark Stevenson$^{2}$\\
  $^{1}$Xi'an Jiaotong Liverpool University\\
  $^{2}$University of Sheffield\\
  \texttt{$^{*}$Haiyang.zhang@xjtlu.edu.cn} }

\begin{document}
\maketitle
\begin{abstract}
Document set expansion aims to identify relevant documents from a large collection based on a small set of documents that are on a fine-grained topic. Previous work shows that PU learning is a promising method for this task. However, some serious issues remain unresolved, i.e. typical challenges that PU methods suffer such as unknown class prior and imbalanced data, and the need for transductive experimental settings. In this paper, we propose a novel  PU learning framework based on density estimation, called puDE, that can handle the above issues. The advantage of puDE is that it neither constrained to the SCAR assumption and nor require any class prior knowledge. We demonstrate the effectiveness of the proposed method using a series of real-world datasets and conclude that our method is a better alternative for the DSE task.  
\end{abstract}

\section{Introduction}
We focus on the task of document set expansion (DSE), as proposed in \cite{EACL2021_DSE}, where a small set of documents belonging to a fine-grained topic is available, and we aim to find all documents on this topic from a large collection. 
This kind of task often arises in the \textit{screening} process of literature curation. In this process, human experts need to carefully go through a larger pool of candidate documents obtained from typical databases to identify relevant documents \cite{neural_ranker2022}.
It is natural for the experts to be aware of the existence of a small set of relevant documents, known as `seed studies',
before initiating the screening process \cite{SIGIR22_seedCollection}. Common methods for modeling this problem involve treating the set of documents as an extended query and making use of information retrieval techniques to re-rank the candidate documents \cite{DocRanking_SIGIR18,ECIR2022_seed_shuai}. However, such methods fail to capture the local or global connections between words and seed studies \cite{EACL2021_DSE}.

\citet{EACL2021_DSE} proposes to model the DSE task as a positive and unlabeled (PU) learning problem and the aim of which is to train a binary classifier with only positive and unlabeled data \cite{survey2020,uPUplessis15,nnPU2017}. 
Their study identifies the challenges of state-of-the-art PU method suffer, such as unknown class prior and imbalanced data. And they propose practical solutions to overcome such challenges.
However, some important issues remain unresolved:
\begin{itemize}
    \item 
    Although the solution introduced in \citet{EACL2021_DSE} shows empirical improvement over typical Non-negative PU (nnPU) \cite{nnPU2017} method, they still fail to address the inherent problems PU methods face except by exploiting the tricks in model implementation, i.e. BER optimization, proportional batching, and use of pre-trained embedding. 
     \item DSE is a transductive problem since we aim to identify all positive documents from the unlabeled set (U). In this case, the unlabeled set should be used for both training and testing. However, in \citet{EACL2021_DSE}, U is split into training, validation, and test sets, with only samples in the test set being classified. Such experimental settings cannot reflect the ground truth performance of the model for the DSE task.
    \item  PU methods that rely on risk estimators, such as nnPU, are based on the assumption of \textit{selected completely at random} (SCAR) labelling mechanism \cite{survey2020}, which assumes that the labeled positive data is identically distributed as the positive data. However, in the DSE scenario, the SCAR assumption does not hold, i.e. the known positive samples is influenced by experts' prior knowledge and selected with bias.
   
\end{itemize}

To address these issues, we propose a novel PU learning framework that is not constrained to the SCAR labeling mechanism assumptions. Similar to SCAR, we focus on the case-control scenario, assuming that the positive and unlabeled data originated from two independent datasets and the unlabeled set is sampled from the real distribution \cite{survey2020}. However, we do not make SCAR's assumption about how the labeled positives data are selected, i.e.,  the data distributions of labeled positives and positives dosen't need to be identical. The primary contributions of this work are: 1) the limitations of the PU solutions proposed in \citet{EACL2021_DSE} for DSE scenario are identified; 2) a new PU learning framework based on the density estimation is proposed, called \textbf{puDE}, that is not constrained to the SCAR assumption and doesn't require any knowledge of class prior; and 3) the proposed method outperform state-of-the-art PU methods on real-world datasets and we conculde that our method is a better solution for the DSE task. 
To our knowledge, this is the first work applying PU methods to transductive  learning.

\section{Preliminary}
In binary classification, we aim to learn a decision function $f:{X}\rightarrow Y$ that can classify $X \in \mathcal{R}^{d}$ into one of the classes $Y \in \{+1, -1\}$. $P(\mathbf{x}, y)$ is the joint probability density of distribution $(X,Y)$ and  $\pi = P(\mathbf{x}|Y=1)$ is the marginal probability density of $\mathbf{x}$ given $Y = 1$, known as class prior. 


The setting of PU, where only a small portion of the positive data is available, is a special case of binary classification \cite{survey2020}. 
The training set ${X}$ can be seen as the combination of the labeled positive set ${X}_{LP}$, the unlabeled positive set ${X}_{UP}$, and the unlabeled negative set ${X}_{UN}$, such that ${X}={X}_{LP} \cup {X}_{UP} \cup {X}_{UN}$. 
Let $s \in \{1,0\}$ be the label status of $y$ ($s = 1$ if labeled, otherwise $s = 0$), there will be:
\begin{equation*}
\small
    \begin{aligned}
    {X}_{LP}&= \{\mathbf{x}|s=1, Y=+1\},\quad
    {X}_{UP} = \{\mathbf{x}|s=0, Y=+1\} \\
    {X}_{UN}&= \{\mathbf{x}|s=0, Y=-1\}, \quad
    {X}_{U}= \{\mathbf{x}|s=0\}
    \end{aligned}
\end{equation*}
The label frequency can be represented as  $c=P(s=1|Y=+1)$ \cite{PU2008Eklan}.

Existing PU models can be divided into two categories based on sampling schemes: censoring scenario and case-control scenario. We here focus on the latter scenario since it is more general than the former one \cite{caseContolNIPS2016}. Case-control scenario assumes a set of positive data and a set of unlabeled data are drawn from $\mathbb{P}(\mathbf{x}|Y=1)$ and $\mathbb{P}(\mathbf{x})$, respectively. In the case-control scenario, the most widely used labeling assumption  is SCAR. It assumes that the positive labeled data are randomly selected from the set of positive data and are identically distributed with the positive unlabeled data, i.e. $ \mathbb{P}(\mathbf{x}|s=1, Y=+1) = \mathbb{P}(\mathbf{x}|s=0, Y=+1)$ \cite{PU2008Eklan}. Many existing PU methods are proposed based on the SCAR assumption and show their efficient for PU problems \cite{uPUplessis15,nnPU2017,vpu2020}. However, as aforementioned, the SCAR hypothesis may not be suitable for many PU applications, \textit{e.g.} in the DSE task, the set of known documents is biased to experts' prior knowledge. 

In this paper, we consider the case-control scenario, i.e. the labeled positive examples are sampled from $\mathbb{P}(\mathbf{x}|Y=1)$, but not the SCAR assumption, i.e. $\mathbb{P}(\mathbf{x}|Y=+1)$ may be different from $ \mathbb{P}(\mathbf{x}|s=1, Y=+1)$. In such case, $\mathbb{P}(\mathbf{x}|Y=+1)$ can be estimated based on the labeled positive examples using density estimation methods without relying on any distributional assumptions.

\section{Proposed Methods} \label{PUL}
\subsection{Task Formulation}
According to \citet{EACL2021_DSE}, we consider the following task: we have a set of labeled positive documents ${X}_{LP}$ on a fine-grained topic and want to find more documents about that topic from a large collection ${X}_{U}$. This task is  essentially the screening prioritisation \cite{SIGIR22_seedCollection} task for literature curation, where some seed studies are known based on the experts's kwnoledge, and the goal is to rank all studies 
in ${X}_{U}$.

\subsection{PU Learning with Density Estimation}
Due to the shortcomings of PU methods that rely on SCAR assumption \cite{EACL2021_DSE}, we propose a new method for PU learning based on the density estimation technique called puDE. It is not constrained to the SCAR labeling mechanism assumptions and do not require any knowledge of the class prior. Given ${X}_{LP}$ and ${X}_{U}$, the objective of puDE is to learn a function to approximate $\mathbb{P}(Y=+1|\mathbf{x})$. According to the Bayesian rule, we have:
\begin{equation}\small
\mathbb{P}(Y=1|X)=\frac{\mathbb{P}(X|Y=1)\mathbb{P}(Y=1)}{\mathbb{P}(X)}
\label{bayesian}
\end{equation}
Let $p: \mathcal{R}^d \rightarrow [0, 1]$ and $q: \mathcal{R}^d \rightarrow [0,1]$ be two trained generative models to approximate the likelihood distributions $\mathbb{P}(X|Y=1)$ and evidence distribution $\mathbb{P}(X)$ respectively. Then, $\mathbb{P}(Y=1|\mathbf{x})$ can be estimated by:
\begin{equation}\small
\vspace{-2mm}
f(\mathbf{x})=\frac{p(\mathbf{x})\cdot\pi}{q(\mathbf{x})}
\label{biCls}
\end{equation}
Note that $\pi$ is a constant prior for each $\mathbf{x}$ which can be ignored in the calculation of $f(\mathbf{x})$ in practice. This is the key advantages of puDE. 
Thus, $f(\mathbf{x})$ can be calculated by estimating $p(\mathbf{x})$ and $q(\mathbf{x})$. Under case-control case, labeled positives are samples from all positive set. Therefore, we can estimate the density of $p(\mathbf{x})$ using samples from ${X}_{LP}$.
In this paper,  we use both nonparametric and parametric density estimation methods to accomplish this task.


\subsubsection{Nonparametric Density Estimation}

Kernel Density Estimation (KDE) is a nonparametric density estimation technique, which has been applied in recommender systems and information retrieval \cite{chakraborty2022kernelIR}. 
An important benefit of KDE is its ability to estimate density without making assumptions about the underlying data distribution. For the given dataset $\{x_1,x_2,\cdots x_n\}$, the estimated density $\hat{f}$ at $x$ is defined as:
    $\widehat{f}_h(x)=\frac{1}{n h} \sum_{i=1}^n K\left(\frac{x-x_i}{h}\right)$
where $h$ is a hyperparameter called bandwidth, and $K$ is a non-negative kernel function. See \citet{Silverman_2018} for more details. 
Using KDE, $f(\mathbf{x})$ is defined as:
\begin{equation}\small
f(\mathbf{x})=\frac{D_p(\mathbf{x})}{D(\mathbf{x})}\pi
\end{equation}
where $D_p$ is the estimated density of positive data which can be estimated by labeled positives ${X}_{LP}$, $D$ is the estimated density of the whole data.


One of the limitations of nonparametric estimation is its tendency to fail in high-dimensional space. 
In practice, we first perform dimension reduction by Variational Autoencoders (VAE) \cite{ReductionVAE2020} to mitigate this issue and then apply the KDE technique to the lower dimensional data.

\subsubsection{Parametric Density Estimation }
The parametric approach used to estimate the density of $p(\mathbf{x})$ and $q(\mathbf{x})$ is the energy-based  model (EBM) \cite{EBM2006}. 
Compared with other parametric density estimation methods, such as VAE \cite{kingma2013auto} and Masked Autoregressive Density Estimators (MADE) \cite{papamakarios2017masked}, EBM do not make any assumption on the form of the probability density they fit. It  aims to learn an energy function which assigns a low energy value to observed data and a high energy value to different values. Based on EBM, $p(\mathbf{x})$ and $q(\mathbf{x})$ in equation \ref{biCls} can be estimated as:
\vspace{-3mm}
\begin{equation}
\small
\begin{aligned}
p_\theta(\mathbf{x})=\frac{e^{-g_{p_{\theta}}(\mathbf{x})}}{Z_{p_{\theta}}}
 \quad \quad 
 q_\theta(\mathbf{x})=\frac{e^{-g_{q_{\theta}}(\mathbf{x})}}{Z_{q_{\theta}}}
\end{aligned}
\end{equation}
where $g_{p_{\theta}}$ and $g_{q_{\theta}}$ represent two parameterized neural networks configured with optimal $\theta$ respectively, $Z_{p_{\theta}}$ and $Z_{q_{\theta}}$ represent the partition functions:
\begin{equation}
\small
Z_{p_\theta}=\int e^{-g_{p_{\theta}}(\mathbf{x})} dx \quad Z_{q_{\theta}}=\int e^{-g_{q_{\theta}}(\mathbf{x})}dx
\end{equation}
Thus, $f(\mathbf{x})$  is rewritten as:
\begin{equation}
\small
\begin{aligned}
f(\mathbf{x}) 
    &=\frac{e^{-g_{p_{\theta}}(\mathbf{x})}}{Z_{p_{\theta}}} / \frac{e^{-g_{q_{\theta}}(\mathbf{x})}}{Z_{q_{\theta}}}\pi \\
    & =e^{(g_{q_{\theta}}(\mathbf{x})-g_{p_{\theta}}(\mathbf{x}))}\left(\frac{Z_{q_{\theta}}}{Z_{p_{\theta}}}\pi\right)
\label{PDE}
\end{aligned}
\end{equation}
where $\frac{Z_{q_{\theta}}}{Z_{p_{\theta}}}\pi$ is a constant for each $\mathbf{x}$ and can be ignored in practice. Hence, $f(\mathbf{x})$ can be approximated by the exponent: $f(\mathbf{x}) := g_{q_{\theta}}(\mathbf{x})-g_{p_{\theta}}(\mathbf{x})$ and the threshold is set to 0 accordingly. 


We employ Markov Chain Monte Carlo (MCMC) sampling to train $g_{q_{\theta}}(\mathbf{x})$ and $g_{p_{\theta}}(\mathbf{x})$,
which aims at optimizing $q_{\theta}(\mathbf{x})$ and $p_{\theta}(\mathbf{x})$, repectively.
The total loss function to minimize is defined as:
\begin{equation}\small
\begin{aligned}
    \alpha\left(-\mathrm{E}_{X_{LP}}\left[\log \mathrm{p}_{\theta}(\mathrm{x})\right]\right)+\beta \left(-\mathrm{E}_{X}\left[\log \mathrm{q}_{\theta}(\mathrm{x})\right]\right) \\ +\gamma\left(R_{\ell_{0-1}}(f(\mathbf{x}))\right)+\mathbb{R}
\end{aligned}
\end{equation}
where $\alpha$, $\beta$ and $\gamma$ are coefficients, $R_{\ell_{0-1}}(f(\mathbf{x}))$ is a normal PU loss, and $\mathbb{R}$ represents the regularization term.

\section{Experiment}
\subsection{ Settings}
We conducted experiments on two distinct groups of datasets. The first group comprised PubMed datasets on 15 fine-grained topics, generated by \cite{EACL2021_DSE} for the DSE task. Within this group, we utilized the document collection focused on three specific topics. Moving on to the second group, it consisted of a single dataset used for Covid-19 study classification \cite{shemilt2022machine}. This dataset serves as a valuable resource for simulating real-world literature curation.

All above datasets were originally designed for inductive classification, where each dataset is split into training, validation and test set. To simulate real-world DSE (transductive case) and for the performance comparison purpose, we treat their test set as ${X}_{U}$ in our setting, and use the test set  ${X}_{U}$ for both training and testing (${X}_{LP}$ and ${X}_{U}$ for training and  ${X}_{U}$ for testing). 
Following \citet{EACL2021_DSE}, the number of labeled positives |LP| is set to $\{20,50\}$ for training on Pubmed datasets. The labeled positives on covid dataset is randomly sampled from their positive training set, and the value of $|LP|$ is set with respect to the ratio of ${X}_{LP}$ over ${X}_{U}$, ranging from 0.01 to 1. The statistics of each set is summarized in Table 1.
\begin{table}[t]
\small
	\centering
	\label{dataset1}
	\begin{tabular}{ccccc}
   \hline
   \text{dataset} &|LP|& $N_{U}$ & $N_{UP}$ & $N_{UN}$ \\
   \hline       
    \multirow{2}{*}{Pubmed-\textit{topic1}}
    & 20& 10012 & 1844 & 8168\\
    & 50& 10027 &2568 & 7459\\
    \multirow{2}{*}{Pubmed-\textit{topic2}}
    &20  & 10012 & 2881 & 7131\\
    &50  & 10027 & 3001 & 7026\\
    \multirow{2}{*}{ Pubmed-\textit{topic3}}
    &20 & 7198 & 1201 & 5997\\
    &50 & 10025 & 1916 & 8109\\
   \text{Covid} & $\{47..4722\}$ &4722  & 2310 & 2412 \\
	
    \hline
	\end{tabular}
 \caption{Statistics of ${X}_{U}$ for each set, where $N_{U}$, $N_{UP}$ and $N_{UN}$, the total number of unlabeled samples, the number of true positive samples and true negatives in the training set. topic 1,2 and 3 correspond to the topics on {Animals+Brain+Rats}, {Adult+Middle Aged+HIV infections} and {Renal Dialysis + Chronic Kidney Failure+ Middle Aged} from \cite{EACL2021_DSE}.}
 \vspace{-4mm}
\end{table}

\subsection{Results}
We use puDE-\textit{kde} and puDE-\textit{em} to denote the proposed PU models that based KDE and EBM, respectively. For puDE-\textit{kde}, the bandwidth is set to 1.9 for both $D_p(x)$ and $D(x)$, and Gaussian function is used as the kernel.  To accommodate the high-dimensional data, we implement a VAE with 256 hidden dimensions and 50 latent dimensions.  For puDE-\textit{em}, the Langevin sampling involves 100 steps with a step size of 0.01. The weights for the total loss function are set as $\alpha=1$, $\beta=1$, and $\gamma=1$. 

nnPU is used as the main competitor in our experiments and 
we implement a transductive version of nnPU using the tricks from \citet{EACL2021_DSE}, denoted as \textit{nnPU-trans}. 
BM25 is act as the baseline for ranking and its results are directly taken from \cite{EACL2021_DSE}. All models, except puDE-\textit{kde}, are implemented using a 6-layer neural network with 200 hidden states per layer. Batch normalization \cite{ioffe2015batch} and leaky ReLU \cite{maas2013rectifier} are applied for each hidden layer. The Adamax optimizer with a learning rate of 1e-3 is employed.

According to Table \ref{tab:f1_comparison}, we can notice that the performance of \textit{nnPU-trans} is much worse than that reported in \cite{EACL2021_DSE} and is similar to BM25, which indicate that the PU solutions proposed in \citet{EACL2021_DSE} is not as effective as they stated for the DSE task. Both puDE methods outperform other methods, with one exception where BM25 get the best result on topic 3 when |LP| equals 20, and show significant improvement over \textit{nnPU-trans}, demonstrating that the proposed PU framework based on density estimation is a better alternative for the DSE task.

Figure 1 demonstrates the F1 results for all model on Covid set, with the ratio of  |LP| over |U| ranging from 0.01 to 1. It can be seen that all models get stable results if more than 10\% of labelled data is avaiable, and both puDE methods consistently shown significant improvements over nnPU with the increase of labeled data. 
\begin{table}[t]
\small
\centering
\begin{tabular}{llp{0.6cm}p{0.7cm}p{0.7cm}p{0.7cm}}
\toprule
|LP| &  Topic &  BM25 & nnPU-trans. & puDE-\textit{kde} & puDE-\textit{em} \\
\midrule
\multirow{3}{*}{20} 
&\textit{topic1}&  32.25 &33.03&37.31 &40.59\\
&\textit{topic2}& 26.75  & 31.30 &36.18 & 39.67\\
&\textit{topic3}& 41.23 & 27.76 &36.63 &35.59\\
\midrule
\multirow{3}{*}{50} 
&\textit{topic1}&  32.80& 38.76 &44.65 & 44.91\\
&\textit{topic2}& 31.85  & 34.16 &44.03 &46.22  \\
&\textit{topic3} & 35.78 &32.84 &36.63 &36.57 \\
\bottomrule
\end{tabular}
  \caption{F1 comparison against baseline and state-of-the-art DES methods.}
    \label{tab:f1_comparison}
\end{table}
\begin{figure}[t]
    \centering
    \includegraphics[width=70mm,height=43mm]{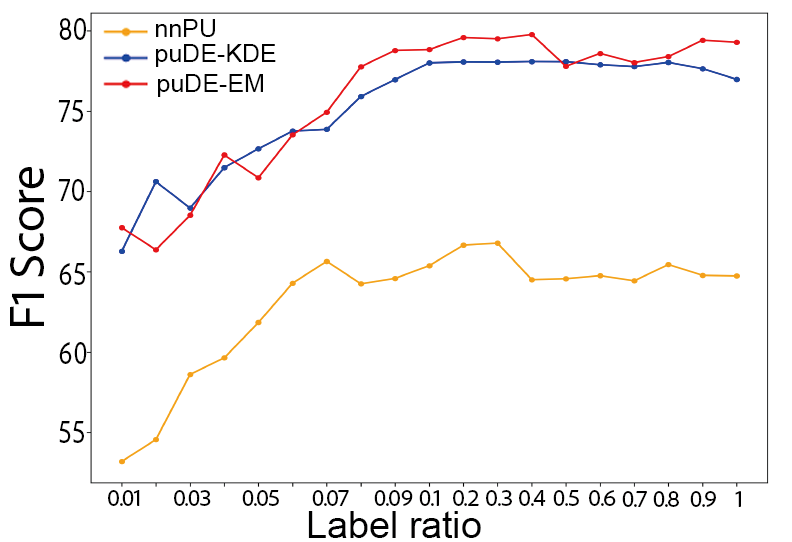}
    \vspace{-2mm}
    \caption{F1 comparison on covid dataset with respect to the ratio of |LP| over |U| ranging from 0.01 to 1.}
    \label{AUC}
    \vspace{-2mm}
\end{figure}

\section{Conclusion}
In this paper, we aim to highlight the limitations of PU solutions propose by \citet{EACL2021_DSE}, and demonstrate that the experimental results in inductive setting cannot be transferred to real-world transductive DSE case. We propose a novel PU learning framework that incorporate two density estimator with the help of Bayesian inference, called puDE. It is not constrained to the SCAR assumption and doesn't require any knowledge of class prior. Empirical experiments verify the effectiveness of our methods. We conclude that our methods is a better solution for the DSE task.


\bibliographystyle{acl_natbib}
\bibliography{emnlp2023-latex/ref}
\end{document}